\documentclass{article}

\newif\ifincludeappendix
\includeappendixtrue

\usepackage{microtype}
\usepackage{graphicx}
\usepackage{amsmath,amssymb}
\usepackage{xcolor}
\usepackage{xurl}

\usepackage{hyperref}
\usepackage[accepted]{mlsys2025}
\makeatletter
\renewcommand{\Notice@String}{}
\makeatother
\hypersetup{pdfsubject={Preprint}}
\newcommand{\correspondingauthornote}{%
  \begingroup
  \renewcommand{\thefootnote}{*}%
  \footnotetext[0]{Corresponding author:
    \href{mailto:akrentsel@berkeley.edu}{akrentsel@berkeley.edu}}%
  \endgroup
}

\mlsystitlerunning{Exo: A Harness Architecture for Runtime Self-Improvement}

\usepackage[english]{babel}
\usepackage{array}
\usepackage{colortbl}
\usepackage{tabularx}
\usepackage{enumitem}
\usepackage{booktabs}
\usepackage{placeins}
\usepackage{pgfplots}
\usepackage{tcolorbox}
\definecolor{exoagent}{RGB}{35,145,64}
\definecolor{exohuman}{RGB}{61,79,190}
\newtcolorbox{exocallout}[2]{
  colback=#1!5!white, colframe=#1, coltext=black,
  boxrule=0.8pt, arc=2pt, boxsep=0pt,
  left=7pt, right=7pt, top=6pt, bottom=6pt,
  before skip=6pt, after skip=6pt,
  fontupper=\small,
  before upper={\textbf{#2}\par\smallskip\noindent}}
\pgfplotsset{compat=1.18}
\usepackage[capitalise,noabbrev]{cleveref}

\crefname{appendix}{Appendix}{Appendices}
\Crefname{appendix}{Appendix}{Appendices}
\crefname{figure}{Fig.}{Figs.}
\Crefname{figure}{Fig.}{Figs.}
\crefname{table}{Tab.}{Tabs.}
\Crefname{table}{Tab.}{Tabs.}
\crefformat{section}{\S#2#1#3}
\crefrangeformat{section}{\S\S#3#1#4--#5#2#6}
\crefmultiformat{section}{\S\S#2#1#3}{ and~#2#1#3}{, #2#1#3}{, and~#2#1#3}
\crefformat{subsection}{\S#2#1#3}
\crefrangeformat{subsection}{\S\S#3#1#4--#5#2#6}
\crefmultiformat{subsection}{\S\S#2#1#3}{ and~#2#1#3}{, #2#1#3}{, and~#2#1#3}
\crefformat{subsubsection}{\S#2#1#3}
\crefrangeformat{subsubsection}{\S\S#3#1#4--#5#2#6}
\crefmultiformat{subsubsection}{\S\S#2#1#3}{ and~#2#1#3}{, #2#1#3}{, and~#2#1#3}

\definecolor{olivegreen}{RGB}{85,107,47}
\newif\ifcomments
\commentsfalse
\ifcomments
    \providecommand{\ak}[1]{{\color{purple}{[AK: #1]}}}
    \providecommand{\martin}[1]{{\color{orange}{[MC: #1]}}}
    \providecommand{\ankur}[1]{{\color{red}{[AG: #1]}}}
    \providecommand{\ion}[1]{{\color{blue}{[ion: #1]}}}
\else
    \providecommand{\ak}[1]{}
    \providecommand{\martin}[1]{}
    \providecommand{\ankur}[1]{}
    \providecommand{\ion}[1]{}
\fi

\newcommand{\topic}[1]{{#1}}
\newcommand{\para}[1]{\par\noindent\textbf{#1}\enspace\ignorespaces}

\newif\ifshowfinaledits
\showfinaleditsfalse

\newcommand{\paperdisplayspacing}{%
  \setlength{\abovedisplayskip}{7pt plus 2pt minus 5pt}%
  \setlength{\belowdisplayskip}{7pt plus 2pt minus 5pt}%
  \setlength{\abovedisplayshortskip}{0pt plus 3pt}%
  \setlength{\belowdisplayshortskip}{4pt plus 3pt minus 3pt}%
}
\AddToHook{env/equation/begin}{\paperdisplayspacing}
\AddToHook{env/equation*/begin}{\paperdisplayspacing}
\AddToHook{env/align/begin}{\paperdisplayspacing}
\AddToHook{env/align*/begin}{\paperdisplayspacing}

\newcommand{\rev}[1]{#1}

\begin{document}
\twocolumn[
\mlsystitle{Exo: A Harness Architecture for Runtime Self-Improvement}

\mlsyssetsymbol{corresponding}{*}
\begin{mlsysauthorlist}
\mlsysauthor{Alexander Krentsel}{berkeley,corresponding}
\mlsysauthor{Sylvia Ratnasamy}{berkeley}
\mlsysauthor{Ion Stoica}{berkeley}
\mlsysauthor{Martin Casado}{a16z}
\mlsysauthor{Ankur Goyal}{braintrust}
\end{mlsysauthorlist}
\mlsysaffiliation{berkeley}{University of California, Berkeley}
\mlsysaffiliation{a16z}{Andreessen Horowitz}
\mlsysaffiliation{braintrust}{Braintrust}
\mlsyscorrespondingauthor{Alexander Krentsel}{akrentsel@berkeley.edu}
\begin{center}
\small
\textsuperscript{1}University of California, Berkeley\quad
\textsuperscript{2}Andreessen Horowitz\quad
\textsuperscript{3}Braintrust
\end{center}

\mlsyskeywords{Agent systems, runtime adaptation, self-improvement}
\vskip 0.3in

\begin{abstract}
Agent harnesses govern how language models act in an environment; designing their policies by hand is difficult, while automated techniques require representative datasets and costly executions. We present \textit{Exo}, an agent runtime architecture for safe learning during ongoing work that collapses the optimization loop into task execution, allowing agents to revise both their harness and their strategies for further improvement. A minimal protected mechanism layer preserves execution history and governs policy activation while the harness evolves through inspectable code changes. On long-running objectives in an open-world game, Exo reduces model cost by 12--53\% relative to Codex and Pi in standalone runs and 45--82\% in transfer runs, while matching task completion rates. Compared with the same starting policy without the ability to edit policy, Exo reduces held-out model cost by 35--61\% on AppWorld and SREGym with comparable correctness, cutting AppWorld's mean task time by 13\% and SREGym's median diagnosis time by 35\%. On Continual Learning Bench coding tasks, model cost including learning falls by 14\%, with both policies resolving every issue.
\end{abstract}
]
\correspondingauthornote

\section{Introduction}
\label{sec:introduction}

\topic{Agents have risen rapidly as a new class of application workload, with wide adoption in developer tools~\cite{claude-code,codex}, personal assistants~\cite{openclaw,hermes}, and enterprise workflows~\cite{pace-insurance,resolve-ai,sierra-siriusxm}.}
These applications combine a language model with a specialized \emph{harness} whose policy constructs context, provides tools, manages memory, and coordinates actions in an environment. Adapting agents to each domain requires adapting these harness policies: a network operations agent needs different tools, context, and execution policies from a coding agent. Harness engineering makes this specialization possible without changing the model's weights, but developers must decide which policies suit each task.

\topic{Building effective harness policies requires substantial development effort.}
Developers must choose prompts, decide what information to retain and how to assemble context, and select which tools and skills to expose. Making these choices by hand is time-consuming and relies on domain expertise and intuition. Agent-building and optimization tools aim to automate this process on benchmarks: Meta-Harness searches over harness implementations~\cite{lee2026metaharness}, GEPA optimizes prompts by reflecting on execution trajectories~\cite{agrawal2025gepa}, AlphaEvolve and SkyDiscover support evaluator-guided program optimization and synthesis~\cite{skydiscover2026,novikov2025alphaevolve}, and industry offers similar solutions~\cite{pace2026tempo}. These systems organize development into an external \textit{optimization loop} that proposes candidate harnesses, executes them on benchmarks, and uses results to guide subsequent revisions.

\topic{However, external optimization depends on evaluation tasks that may be unavailable and repeated executions that are increasingly costly, while leaving harness policies fixed during subsequent work.}
In new domains, benchmark tasks and faithful evaluation environments are often unavailable; one study reports that 75\% of interviewed teams evaluate their agents without benchmark sets~\cite{pan2025map}. Even when benchmarks exist, the trend toward longer task horizons increases the cost of evaluating candidates through task completion, with some reported executions already lasting days or weeks~\cite{openai2025codexmax,lin2026selfdriving,carlini2026compiler}. Benchmarks can also become outdated as the environment changes. Finally, the resulting harness policies remain fixed during execution, even when an agent's runtime experience may expose a possible gap. Taken together, these challenges can make external agent-building techniques impractical in real-world settings or limit their effectiveness.

\begin{figure}[t]
  \centering
  \includegraphics[width=\columnwidth]{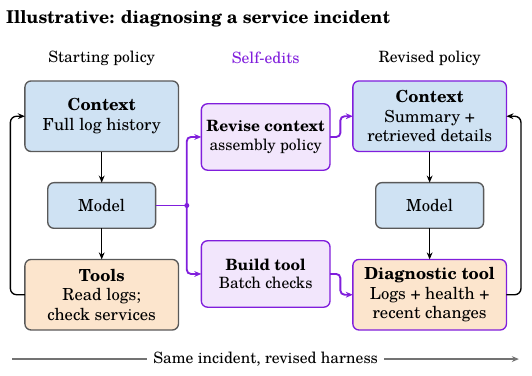}
  \caption{Illustrative harness adaptation during a service incident. The agent builds a diagnostic tool and revises context assembly while continuing the same task. Model weights remain unchanged.}
  \label{fig:collapsing-the-loop}
\end{figure}

\topic{In this work, we propose an alternative approach to harness adaptation: equip a minimal agent to inspect and edit its own harness, then let it use experience from ongoing work to revise and adapt itself at runtime.}
As we explain in \cref{sec:approach-collapse}, two properties enable this departure from traditional external adaptive control: the agent can inspect and reason about the code implementing its own harness, and it can implement policy revisions with the same code-editing tools it uses for ordinary task actions. This enables us to \emph{collapse the optimization loop}: the agent chooses both task actions and changes to how it acts (\cref{fig:collapsing-the-loop}). Adaptation need not wait for retrospective review of completed execution traces: with self-editing available as an explicit action, the agent can explore alternative behaviors, assess their effects, and decide whether to incorporate them into its harness during the task.
This design allows for the very policy the agent uses to govern its action to evolve at runtime with experience, including to improve its ability to improve itself. 

\topic{However, enabling agents to edit themselves introduces two core systems challenges.}
First, the resulting \emph{fate-sharing} between the optimizer and the running system means that a faulty edit can disable both the agent and the logic needed to diagnose and repair it.
Second, self-edits can erase unprotected state – execution history, memory, and task artifacts – destroying accumulated progress and the record of past actions, successes, and failures. An agent cannot repair itself if it cannot execute, nor learn from experience it has lost.

\topic{To address these, we present Exo, an agent runtime architecture for learning and adaptation during ongoing work, with mechanisms to support safe evolution of an agent's harness policy.}
Our key idea is to partition the agent into an evolvable policy layer and a minimal set of protected mechanisms necessary for safe operation. As we show in \cref{sec:design}, these mechanisms preserve execution history and prior state across policy revisions and keep recovery authority outside the editable layer. Inspired by operating systems work on separating policy from mechanism, particularly exokernels~\cite{engler1995exokernel}, we expose primitive interfaces that leave as much policy as possible in the evolvable layer, including how the agent uses its experience and improves itself.
Exo's architecture makes learning observable and adaptable while limiting the damage a faulty revision can cause.
The agent can examine its execution history, including outcomes, model costs, and cache use, to identify improvements grounded in the environment where it actually works. It can use that evidence to revise both its task policy and its optimization strategy. Throughout this process, protected history and snapshots preserve the experience and recovery points needed to continue learning after a failed revision, while an independent guardian controls when revised harness policies take effect (\cref{sec:design}). The resulting adaptations are interpretable: they appear as explicit changes to harness code that developers can inspect.

\topic{We evaluate two benefits of runtime learning: improving efficiency within complex tasks and transferring the resulting specialization to similar tasks.}
We study long-task adaptation in RuneScape, a dynamic, open-world multiplayer role-playing game, across combat, crafting, and currency objectives. In standalone runs, Exo completes combat and crafting at 12--53\% lower model cost than Codex and Pi and earns 44--125\% more currency per model dollar. In combat transfer runs on fresh accounts, the cost reduction is 72--82\%. We also compare Exo to the same starting policy without the ability to edit policy on AppWorld, SREGym, and Continual Learning Bench (CLB) coding tasks~\cite{asawa2026continuallearningbench}. On AppWorld, Exo reduces held-out cost by 61\% at the same observed 96.9\% pass rate and total cost including learning by 43\%. On 20 held-out SREGym incidents, it reduces median diagnosis time and test cost by 35\%, and model rounds by 38\%. On CLB, it uses 15\% fewer shell steps and 14\% less model cost including reflection, with matching issue resolution rates across five runs.

\topic{In this work, we make three contributions:}
\begin{itemize}[leftmargin=*,topsep=0pt,itemsep=0pt,parsep=0pt]
    \item We propose runtime harness self-editing and identify the properties that allow the running agent to revise both its task policy and its improvement strategy, collapsing the optimization loop.
    \item We identify the challenges of runtime self-editing and address them through an architecture that separates evolvable policy from protected mechanisms for preserving state and controlling policy changes.
    \item We implement and release Exo as open-source software~\cite{exo2026}, and evaluate it across open-world objectives and three graded benchmarks, demonstrating efficiency gains during ongoing work and reuse of learned policies on subsequent tasks.
\end{itemize}

\section{Background}
\label{sec:background}

We first define agents and their policy, then describe the challenges of adapting policies to real-world workloads.

\subsection{Agents and Harnesses}
\label{sec:background-harnesses}

An \emph{agent} couples language-model inference to an environment through a repeated interaction loop~\cite{yao2023react}. We view the \emph{harness} as the scaffolding surrounding the model that (1) assembles the model's input and (2) executes and observes actions on the \textit{environment}~\cite{anthropic2025contextengineering}, which is most often a computing sandbox. A \emph{task} is a unit of work with an objective and completion criteria, which may be implicit. Task granularity varies: an SRE agent might manage one software rollout or a month's availability SLO across all rollouts.

\para{Harness policy.}
We use \emph{policy} to refer to all of the programmable rules and procedures governing how an agent uses its model, state, and environment. These include tool implementations and selection; prompting; context assembly and compaction; memory and skill indexing and retrieval; and external integrations, such as service adapters and task scheduling. These choices are encoded in harness code, prompts, configuration, and reusable skills. Policy also includes the strategy for deciding when and how to attempt further improvements.

Though tools in policy and agent-written code in the environment are similar in spirit (in what they can accomplish), they differ in how they are managed by the agent. Policy-level tools are often wired directly into context assembly, whereas code in the environment must be remembered and managed entirely by model reasoning.

\subsection{Challenges in Harness Adaptation}
\label{sec:background-challenges}

Automated optimization methods develop harness policies through an external loop that proposes candidates, executes them on evaluation tasks, and uses the results to guide revisions~\cite{agrawal2025gepa,lee2026metaharness}. Three limitations motivate a different approach.

\para{Evaluation datasets are difficult to obtain and maintain.}
New domains often lack representative task sets and faithful evaluation environments. In a study of production agents, 75\% of interviewed teams evaluated without benchmark sets~\cite{pan2025map}. Constructing a benchmark requires capturing relevant tasks, expected outcomes, and the environment in which actions take effect. Even then, changing services, interfaces, and user requirements can make those tasks and environments outdated~\cite{krentsel2026twogaps}. An optimizer may consequently improve performance against a proxy that no longer reflects reality.

\para{Longer tasks increase evaluation cost.}
Evaluated task horizons are increasing~\cite{metr2026horizons}, with software-building and theorem-proving projects already spanning weeks~\cite{carlini2026compiler, openai2026navierstokes}. As executions lengthen, evaluating successive candidate policies through complete task runs becomes more expensive. Waiting for task completion also delays feedback, even when an inefficient procedure or recurring failure becomes apparent much earlier.

\para{Fixed harness policies limit runtime adaptation.}
Even a policy optimized for a representative workload can encounter new patterns during execution. A fixed harness can react to observations and retain memories, but its procedures for constructing context and executing actions remain unchanged. For example, an agent may recognize that most diagnostic output is irrelevant while its context-construction policy continues to include it. Without a path from that observation to a policy revision, the same inefficiency can recur throughout the task.

\section{Approach}
\label{sec:state-and-self-editing}

\begin{figure}[t]
  \centering
  \includegraphics[width=0.7\columnwidth]{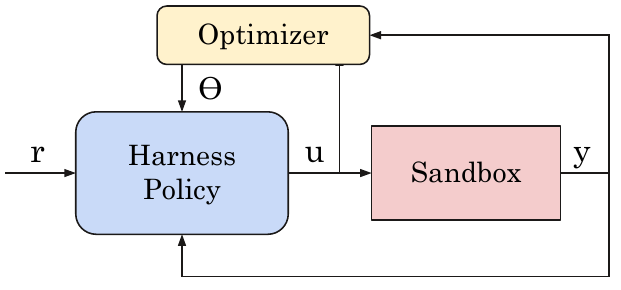}
  \par\vspace{0.25em}
  \makebox[\columnwidth][l]{\small\textbf{(a)} Classical "adaptive control" with optimizer and controller.}
  \par\vspace{0.75em}
  \includegraphics[width=0.7\columnwidth]{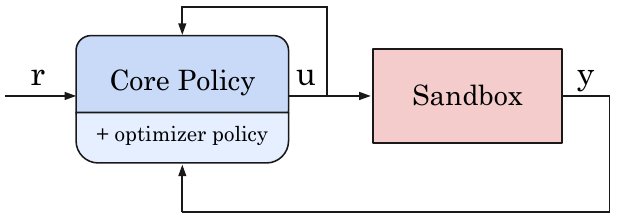}
  \par\vspace{0.25em}
  \makebox[\columnwidth][l]{\small\textbf{(b)} Collapsed optimizer + controller (Exo's approach).}
  \caption{Collapsing an adaptive optimization loop. In (a), a separate
  optimizer maps execution feedback to harness updates $\theta$. In (b), the agent chooses task and revision actions through the same action interface
  $u$.}
  \label{fig:adaptive-loop-collapse}
\end{figure}

To address the challenges raised in \cref{sec:background}, we propose an alternate method of policy creation: runtime harness adaptation, where the running agent uses experience from ongoing work to revise the policy governing that work. This brings task actions and policy revisions into the same decision process.

\subsection{Collapsing the Optimization Loop}
\label{sec:approach-collapse}

The external optimization systems described in \cref{sec:background-challenges} can be viewed through the lens of control theory as adaptive control loops~\cite{astrom2008feedbacksystems}. In a traditional adaptive control model, shown in \cref{fig:adaptive-loop-collapse}(a), the controller (harness policy) maps an objective $r$ and observations of the environment $y$ to task actions $u$ (tool calls). An optimizer then observes the same execution and produces an update $\theta$ to the controller policy.

Harnesses have two special properties that allow us to bring the optimizer into the core policy itself. First, a harness revision can be realized in the same action domain as ordinary task work, which we refer to as \textit{optimizer symmetry}. More precisely, the update $\theta$ can be produced by a sequence of actions in $\mathcal{U}^*$: the agent can inspect source, edit files, run tests, and request activation using tools already exposed through $u$. Revision actions therefore become choices available to the core policy alongside actions that act directly on the task.

Second, the harness is \emph{operationally homoiconic}: its policy is code, and that code is represented as files the agent can observe and manipulate with its code-editing tools. The object being optimized is consequently available as data in the optimizer's environment. Together, same-domain actions and editable policy representation allow the optimizer's proposal logic to move inside the harness, producing the collapsed loop in \cref{fig:adaptive-loop-collapse}(b).

Because the policy governing when and how to revise the harness is itself editable, the agent can revise both its task behavior and its strategy for further improvement. 
Each successor can adopt a new policy while retaining the ability to revise itself further, independent of model retraining.

\subsection{State and Replacement Contracts}
\label{sec:approach-contracts}
\label{sec:approach-requirements}

Bringing revision into ongoing execution creates two systems challenges. First, it introduces \emph{fate-sharing} between core logic and optimizer logic, meaning a faulty edit can disable both the agent and the logic it would use to diagnose and repair itself. Second, edits to \emph{unprotected state} can erase accumulated progress and execution evidence, including records of failed revisions.

To address these two critical failure modes, we turn to system architecture. We first model the system's runtime state and transitions, then specify the properties the architecture must enforce as the agent revises its policy.

\para{State and transitions.}
We call the replaceable program that implements the agent's harness policy the \emph{executor}. We take runtime state at step $t$ to be $s_t$, and let a protected mechanism $K$ apply the executor's chosen operation $a_t$:
\begin{align}
    s_t &= (d_t, e_t, \pi_t), \label{eq:exo-state}\\
    s_{t+1} &= K(s_t, a_t).
    \label{eq:exo-transition}
\end{align}
Here $d_t$ contains durable data, including history, memory, and artifacts; $e_t$ contains sandbox files, dependencies, and processes; and $\pi_t$ identifies the active executor implementation and configuration.
Operations may update data, act in the sandbox, or request replacement. In order to stay protected, the storage and replacement rules in $K$ must remain outside the executor's editing authority.

\para{Storage contract.} Let $H(d)$ denote canonical execution history: the record of messages, tool calls and results, and policy-replacement events. The storage contract requires this history to be append-only:
\begin{equation}
    H(d_t) \preceq H(d_{t+1}),
    \label{eq:exo-history}
\end{equation}
where $\preceq$ denotes the prefix relation. Existing records cannot be overwritten or deleted. Memory and task artifacts may have mutable current values, but updates must retain prior versions for recovery. These rules preserve evidence without prescribing how the policy uses it.

\para{Replacement contract.}
Editing source produces an arbitrary candidate $\pi'$, which may have any sort of behavior including crashing; activating this as the active policy must therefore require a check that lies outside of the policy itself. We call this mechanism the \textit{guardian} $G(s_t,\pi')$, which evaluates to $\{0,1\}$. Thus, for a replacement request,
\begin{equation}
    \pi_{t+1} =
    \begin{cases}
        \pi', & \text{if } G(s_t,\pi') = 1,\\
        \pi_t, & \text{otherwise}.
    \end{cases}
    \label{eq:exo-replacement}
\end{equation}

Policy replacement preserves durable data and leaves the sandbox unchanged, while appending replacement events to history; the acceptance decision may inspect both.

Taken together, these contracts enable the agent to revise its policy with maximal flexibility while ensuring that edits cannot erase protected history or bypass replacement checks. Notably, this protection holds regardless of how the agent chooses its edits, but does not on its own guarantee improved behavior or continued progress. \rev{We give the preservation proof and discuss these guarantees' assumptions and limits in \cref{supp-app:contracts}.}

\section{Design}
\label{sec:design}

\begin{figure*}[t]
\centering\color{black}
\begin{minipage}[b]{0.66\textwidth}
\centering
\includegraphics[width=\linewidth]{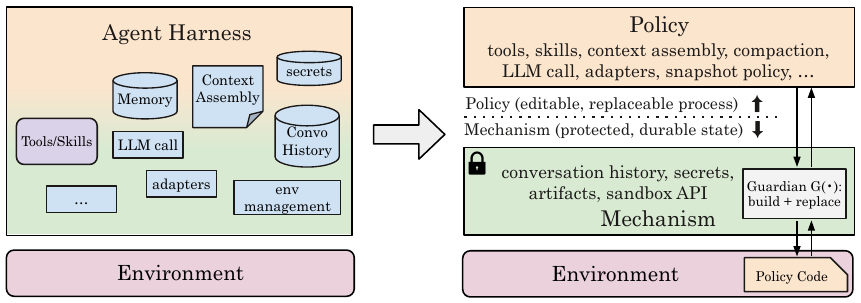}
\par\smallskip
{\small\textbf{(a)} Policy--mechanism partition.}
\end{minipage}\hfill
\begin{minipage}[b]{0.32\textwidth}
\centering
\includegraphics[width=\linewidth]{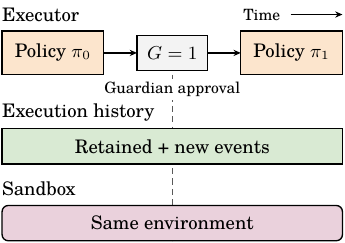}
\par\smallskip
{\small\textbf{(b)} State across replacement.}
\end{minipage}
\caption{Exo separates editable policy from protected mechanisms (a). Policy source is exposed for self-editing, with executor activation controlled by a guardian. Across an accepted executor replacement (b), history is retained and extended, and the sandbox can remain in place. State and replacement contracts are given in \cref{sec:approach-contracts}.}
\label{fig:design-architecture}
\end{figure*}

The goal of Exo's architecture is to implement the state and replacement contracts in \cref{sec:approach-contracts} to provide safety, while enabling the agent to self-edit at runtime with as much freedom as possible to change how it works. We now describe our architectural design decisions to enable this.

\subsection{Partitioning Policy and Mechanism}
\label{sec:design-architecture}
\label{sec:design-executor}

We consider agent architecture through the lens of a computing system. The traditional view of harnesses combines two different concerns: the infrastructure required to serve an agent, including durable history, privileged resources, and sandboxes; and the policies governing how the agent thinks and acts, including prompt assembly, compaction, memory, and tools. This makes self-evolution difficult to do safely, as the evolvable policy is not distinct from protected resources. 

Our key guiding design decision is to partition the harness into two distinct layers: a protected mechanism layer and an editable policy layer, akin to the kernel and userspace split in operating system (OS) design. To determine what functionality should go in each, we draw on inspiration from Exokernel~\cite{engler1995exokernel}, aiming to isolate the minimal set of functionality needed for our safety properties, and to expose it at the lowest level of abstraction to maximize the space of behaviors that can be expressed above it.

We show our chosen partition in \cref{fig:design-architecture}(a). Below the boundary, Exo protects the state and mechanisms needed to preserve continuity across self-edits. These are the canonical execution history, which is exposed through an append-only queryable interface annotated with costs, along with a versioned artifact store that can hold arbitrary data  such as memory, skills, or sandbox snapshots. Finally, Exo provides the protected mechanisms for credential storage, sandbox API, and the independent guardian mechanism that controls activation of replacement executors.

All other harness functionality is expressed in the policy layer, which we call the \textit{executor}. These include tool implementations and selection; context assembly and compaction; prompting; memory and skill indexing and retrieval; and external integrations, including service adapters and task scheduling.
For example, the protected mechanism preserves execution history, while the policy decides which parts of that history are included in the prompt sent to the model. The agent can revise any policy code above the protected boundary, including its tools, model calls, compaction, and the procedures governing when and how it attempts further improvements.

\subsection{Enabling Runtime Self-Editing}
\label{sec:design-replacement}

To enable runtime self-editing on this architecture, we take two steps:
\begingroup
\setlength{\parskip}{0pt} %
\begin{enumerate}[label=(\arabic*),wide=0pt,nosep,labelsep=0.4em]
\item Mount the executor code in the sandbox, allowing the agent to inspect and edit it with its existing tools.
\item Provide a tool to request executor replacement through a protected guardian, which builds the candidate and activates it only if it passes a guardian function $G$.
\end{enumerate}

The agent can thus inspect and edit its own implementation during task execution, then request activation of the revised policy (\cref{fig:design-architecture}(b)). The guardian runs independently of the executor, and replacement requests and outcomes are recorded in the event history. Our default guardian checks build success; richer implementations could evaluate candidate behavior over a longer observation period (\cref{sec:discussion-guardians}).
\endgroup

\para{Independent rollback.}
\label{sec:design-state}
\label{sec:design-sandboxes}
Exo's split also enables independent rollback of executor policy, durable agent state, and environment sandboxes. Executor rollback restores an earlier code version and reactivates it through the guardian. Append-only history supports forking from an earlier event, while versioned artifacts retain earlier state for recovery. The sandbox API separately supports snapshotting and restoring the task environment. These operations preserve the original execution record, allowing the agent to retain evidence of what it tried even after a rollback. Snapshot scope and limits on reversing external effects are discussed in \cref{supp-app:contracts}.

\subsection{Internal Legibility}
\label{sec:design-legibility}

To guide self-improvement, an agent needs to inspect both its past behavior and the costs it incurred. Exo exposes execution history for retrieval and analysis, annotating model-message events with cost and token counts, including cached tokens. This makes earlier execution available for inspection even when it is no longer included in the model's context. Policy code can surface these annotations alongside the messages it retrieves to assemble context, or through tools that summarize execution statistics. Our implementation provides a reporting tool covering spending, cached input share, repeated tool calls, and context composition. For example, the agent can inspect whether a context revision reduces input tokens while preserving cache reuse.

\para{Implementation and deployment.}
We have implemented and released Exo as open-source software~\cite{exo2026}. The core runtime architecture, including protected mechanisms, totaled $\sim$20,000 lines of code, and the base starting policy comprises $\sim$1000 lines. The runtime is deployed in production, having served O(10k) diverse tasks over O(5m) messages.

\begin{figure*}[t]
\centering
\includegraphics[width=0.9\textwidth]{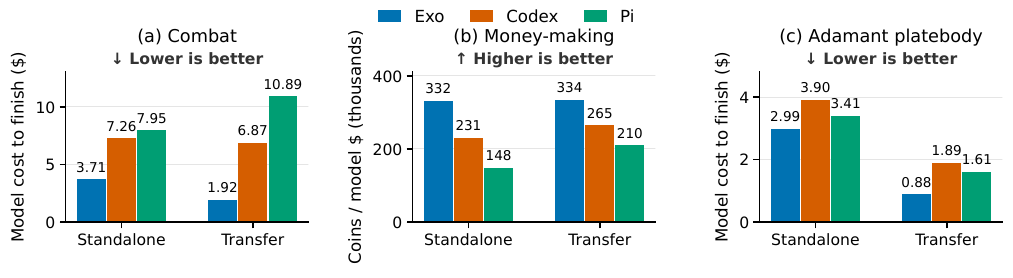}
\caption{Open-world outcomes in standalone and transfer runs. Combat and crafting report model cost to finish (lower is better); money-making reports coins per model dollar near eight active hours (higher is better).}
\label{fig:eval-efficiency}
\end{figure*}

\section{Evaluation}
\label{sec:evaluation}

We evaluate two facets of Exo's runtime learning: (1) whether specialization improves efficiency within an ongoing task, and (2) whether the resulting tools and policies remain useful on subsequent tasks. We first compare Exo with existing harnesses on long-running open-world objectives, then examine how specialization develops and transfers. Finally, we evaluate efficiency and correctness across graded application, infrastructure, and coding tasks.

\subsection{Experimental Setup}
\label{sec:eval-setup}

\para{Comparisons.}
Our open-world experiments compare Exo, Codex~\cite{openai2026codexcli}, and Pi~\cite{zechner2026pi} using \texttt{gpt-6.1-sol} at medium reasoning effort across different objectives. For AppWorld, SREGym, and Continual Learning Bench (CLB) coding, we use \texttt{gpt-5.6-sol} and compare Exo to the same starting policy without the ability to edit policy. This control includes standard tools for coding, shell access, subtask planning, and memory and is labeled \emph{Control} in our figures. We describe each workload's task sequence and feedback below.

\para{Metrics.}
We report task outcomes alongside model cost and time. For finite game objectives, we measure cost to completion and progress per model dollar over a fixed window. All costs use common token pricing that accounts for caching. For graded tasks, we report correctness and execution efficiency, distinguishing cost on held-out tasks from total cost including learning and reflection. A \emph{standalone run} starts without retained agent state; for game objectives, a \emph{transfer run} repeats the objective on a fresh account while retaining agent state.

\subsection{Specialization During Open-World Execution}
\label{sec:eval-open-world}

We first examine runtime specialization in a long-running task in an emulated environment based on RuneScape, an open-world multiplayer role-playing game~\cite{runescape}. Agents act through the common RS-SDK interface~\cite{bittker2026rssdk} to pursue objectives, starting from new accounts. We select three distinct objectives: training all combat stats to their maximum; accumulating currency through any available strategy; and directly crafting a challenging item, the adamant platebody, which requires resource gathering and training, without buying the finished item. These objectives require hours of coordinated action across a very large open world with a dynamic environment of monsters and shared resources. Combat runs were performed on a public server with 1,200-1,500 other players and bots. The other tasks reported were run on a shared private server due to public server overload.
All agents run concurrently in the same game world.

\textbf{Exo completes the finite objectives at lower model cost than Codex and Pi, and earns more currency per model dollar.}
\Cref{fig:eval-efficiency} compares standalone and transfer outcomes; \cref{supp-tab:eval-open-world} provides a tabular view of these outcomes. In standalone combat, Exo spends \$3.71, versus \$7.26 for Codex and \$7.95 for Pi; in crafting, it spends \$2.99, versus \$3.90 and \$3.41. Together these yield 12--53\% lower completion cost. For money-making, Exo earns approximately 332k coins per dollar, versus 231k and 148k, an improvement of 44--125\%.
All arms finish the finite objectives. 

\subsubsection{Looking Inside Runtime Specialization}
\label{sec:eval-combat}

\begin{figure*}[t]
\centering
\includegraphics[width=0.8\textwidth]{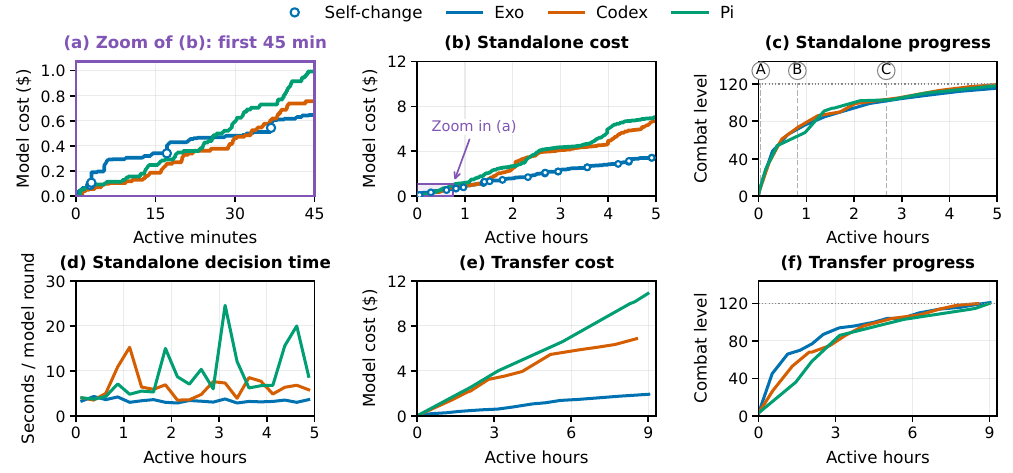}
\caption{Combat specialization and reuse. (a) First 45 active minutes, magnifying the purple box in (b). Circles in (a,b) mark Exo's tool changes and memory updates. (b--d) Standalone cost, progress, and decision time through five active hours; A--C identify tool changes discussed in the text. (e,f) Transfer cost and progress on fresh accounts with prior agent state retained; A spike in public-server usage lengthens transfer runs.
Cost in (a,b) steps at model calls. Decision time is the median time from input to the next model action in each 15-minute window, including provider and network delay.}
\label{fig:eval-combat}
\end{figure*}

\textbf{Exo shows an early investment in specialization followed by a lower rate of model spending during the same task.}
\cref{fig:eval-combat}(a) shows progress trajectories across agents for the combat task. At ten minutes, Exo has spent approximately \$0.31, versus \$0.12 for Codex and \$0.16 for Pi (\cref{fig:eval-combat}(a)). Much of the early work for Exo builds and revises a reusable combat tool and edits context assembly policy, while Codex and Pi make progress towards their goals with comparatively less reasoning. However, as a result of these changes, Exo quickly optimizes its spend rate, with Pi's cumulative spending overtaking Exo at 25 minutes, and Codex's at 86 minutes. We call this early expense the "specialization cost", where self-improvement pays an early premium for improved later performance within the standalone execution.

\textbf{Exo's specialization appears as concrete changes to its tools, action procedures, and context policy.}
\Cref{fig:eval-combat}(c) highlights three of the sixteen self-edits Exo made at runtime. At A, Exo creates \texttt{train\_combat}, which batches melee actions and reports experience gained and observations. After repeated failures to attack a target, it revises the tool at B to skip stalled targets and retreat earlier. At C, it builds \texttt{combat\_cycle}, which selects the lowest uncapped melee skill, trains for 180 seconds, collects nearby resources, and stops on low health. Later, it changes context assembly to retain only four earlier turn summaries, leaving older strategy in durable memory. These edits replace repeated model decisions with reusable procedures and change how the agent presents its experience to the model, which result in meaningfully more efficient agent behavior.

\textbf{Specialization may not repay its cost when work offers little reuse.}
We selected one additional objective representative of the least repetitive class of task: completing as many quests as possible. Each quest is completed once and requires largely distinct steps, items, and locations. Across three runs per arm, Exo averages \$0.28 per quest point, versus \$0.24 for Codex and \$0.21 for Pi. Exo builds reusable dialogue tools, but the quest-specific work leaves little opportunity to amortize specialization cost. Thus other harnesses without self-reflection or improvement operate more cheaply for such a task.

\subsection{Transfer of the Specialized Harness}
\label{sec:eval-transfer}

\textbf{The tools and policies learned during standalone runs remain useful when the agent repeats the objectives on fresh accounts.}
All three arms retain their history and sandbox state; Exo also carries its self-modifications. As shown in \cref{fig:eval-efficiency}, Exo's combat cost falls from \$3.71 to \$1.92, versus transfer costs of \$6.87 for Codex and \$10.89 for Pi. Its crafting cost falls from \$2.99 to \$0.88, versus \$1.89 and \$1.61. Exo thus completes these transfer objectives at 45--82\% lower model cost than the other harnesses. Its own efficiency improves from the standalone to transfer run as it has paid the upfront learning cost. Its money-making efficiency remains approximately unchanged, as it found the close-to-optimal strategy early in the standalone run.

\textbf{In combat transfer, Exo reuses its specialized harness without further self-changes.}
The low spending rate in \cref{fig:eval-combat}(e) reflects the application of tools and context changes acquired during the preceding run. All arms retain their experience, but their resulting policies use that experience differently: Exo can invoke its learned action procedures directly. As \cref{fig:eval-combat}(f) shows, all three arms reach the combat target after approximately nine active hours, longer than the standalone run, while Exo maintains its model-cost advantage. This experiment demonstrates reuse on a repeated objective under changed execution conditions.

\begin{figure*}[t]
\begin{minipage}[t]{0.48\textwidth}
\vspace{0pt}\centering
\includegraphics[width=\linewidth]{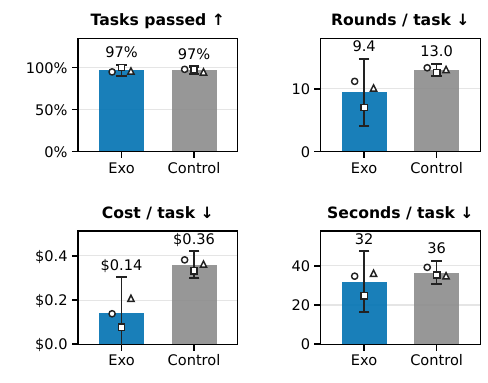}
\caption{AppWorld held-out outcomes across three seeds. Bars are means, symbols show seeds, and error bars are 95\% $t$-intervals. Control uses the same starting policy without the ability to edit policy.}
\label{fig:eval-appworld-summary}
\end{minipage}\hfill
\begin{minipage}[t]{0.48\textwidth}
\vspace{0pt}\centering
\includegraphics[width=\linewidth]{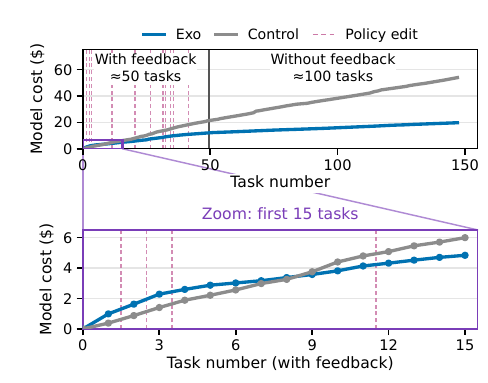}
\caption{AppWorld: cumulative model cost including reflection over a full run (top), with the purple box enlarged below. Dashed lines mark Exo's policy edits. Exo's cumulative cost falls below Control's at task 9.}
\label{fig:eval-appworld-example}
\end{minipage}
\end{figure*}

\subsection{Learning Across Graded Tasks}
\label{sec:eval-cl}

We next evaluate whether harness adaptation improves efficiency across application, infrastructure, and coding tasks. For all three benchmarks, we compare against the same starting policy without the ability to edit policy. We restrict network access on all arms to prevent cheating. All arms receive the same correctness feedback after learning tasks. AppWorld and SREGym then use held-out tasks without feedback; CLB measures learning across a fixed issue sequence. \Cref{tab:eval-benchmarks} summarizes the primary outcomes.

\begin{table}[t]
\centering\normalsize
\setlength{\tabcolsep}{5pt}
\renewcommand{\arraystretch}{1.12}
\definecolor{benchmarkgain}{HTML}{D8EBD3}
\caption{Graded-task outcomes for Exo vs identical policy without policy editing (Control). Arrows indicate the preferred direction. AppWorld and CLB values are means; SREGym reports median time and rounds and mean cost. CLB costs include reflection.}
\label{tab:eval-benchmarks}
\begin{tabularx}{\columnwidth}{@{}>{\raggedright\arraybackslash}X r >{\columncolor{white}[\tabcolsep][0pt]\bfseries}r@{}}
\toprule
\textbf{Metric} & \textbf{Control} & Exo \\
\midrule
\rowcolor{black!6}[0pt][0pt]
\multicolumn{3}{@{}l@{}}{\strut\textbf{AppWorld}\quad\small Held-out tasks; 3 seeds} \\
Tasks passed $\uparrow$ & \cellcolor{black!4}\textbf{96.9\%} & \cellcolor{black!4}96.9\% \\
Model cost / task $\downarrow$ & \$0.359 & \cellcolor{benchmarkgain}\$0.141 \\
Time / task $\downarrow$ & 36.5\,s & \cellcolor{benchmarkgain}31.9\,s \\
\addlinespace[4pt]
\rowcolor{black!6}[0pt][0pt]
\multicolumn{3}{@{}l@{}}{\strut\textbf{SREGym}\quad\small 20 held-out incidents} \\
Diagnosis time $\downarrow$ & 81\,s & \cellcolor{benchmarkgain}53\,s \\
Model rounds / incident $\downarrow$ & 16 & \cellcolor{benchmarkgain}10 \\
Model cost / incident $\downarrow$ & \$0.892 & \cellcolor{benchmarkgain}\$0.584 \\
\addlinespace[4pt]
\rowcolor{black!6}[0pt][0pt]
\multicolumn{3}{@{}l@{}}{\strut\textbf{CLB coding}\quad\small All tasks; 5 runs} \\
Issues resolved $\uparrow$ & \cellcolor{black!4}\textbf{100\%} & \cellcolor{black!4}100\% \\
Shell steps / issue $\downarrow$ & 7.42 & \cellcolor{benchmarkgain}6.28 \\
Model cost / issue $\downarrow$ & \$0.507 & \cellcolor{benchmarkgain}\$0.436 \\
\bottomrule
\end{tabularx}
\end{table}

\subsubsection{AppWorld}
\label{sec:eval-appworld}

AppWorld~\cite{trivedi2024appworld} presents user requests spanning nine everyday applications, such as computing payments made to roommates. Exo learns on one variant of each of 50 scenarios, with grader feedback and reflection after each task, then executes 100 held-out variants of those scenarios without feedback or reflection. We repeat this sequence with three seeds, using the same task order for both arms and all seeds.

\textbf{Exo reduces held-out model cost by 61\% at the same observed pass rate, with savings that also cover the learning period.}
Both arms pass 96.9\% of held-out tasks. Exo costs \$0.141 per task versus \$0.359, uses 27\% fewer model rounds, and reduces mean task time by 13\% (\cref{fig:eval-appworld-summary}). Including learning and reflection, mean total cost falls from \$57.36 to \$32.70, a 43\% reduction.

\textbf{Inspection of the learned harness shows reusable API-documentation tools and changes to context construction.}
For example, \texttt{appworld\_api\_docs} fetches compact documentation for several relevant APIs in one call, avoiding repeated discovery. Exo also moves its changing task list out of the standing prompt and revises context construction; cached input share rises from 57.5\% to 86.1\%. \Cref{fig:eval-appworld-example} traces seed 2: Exo revises its harness after each of the first three tasks, and its cumulative cost falls below Control's at task 9. The learned procedures differ across seeds. Seeds 1 and 2 build broadly reusable tools, while seed 3 creates many scenario-specific tools and spends more during learning. All three nevertheless achieve lower held-out cost. This variation shows that specialization need not converge to a single implementation.

\subsubsection{SREGym}
\label{sec:eval-sre}

SREGym~\cite{clark2026sregym} evaluates agents diagnosing and mitigating faults injected into running services. The efficiency comparison uses 21 learning incidents with feedback and reflection, followed by 20 unseen incidents from the same fault families on different applications, without feedback or reflection.

\textbf{Exo's learned policy reduces diagnosis time, model rounds, and the cost of handling subsequent incidents.}
Median diagnosis time falls from 81 to 53 seconds (35\%), median model rounds from 16 to 10 (38\%), and mean model cost per held-out incident from \$0.892 to \$0.584 (35\%; \cref{tab:eval-benchmarks}). Correctness is comparable across additional holdout runs.

\textbf{Exo builds and repeatedly refines a diagnostic tool that packages evidence needed across incidents.}
Exo creates and refines an incident-snapshot tool that collects unhealthy pods, revision and template anomalies, and suspicious log lines, revising this tool five times and reusing it as its first diagnostic step on held-out incidents. The tool consolidates evidence that otherwise requires multiple shell calls, giving subsequent diagnoses a reusable starting procedure. It builds an additional 5 tools for targeted investigation.

\subsubsection{Continual Learning Bench Coding}
\label{sec:eval-clb}

\textbf{On CLB coding tasks, Exo uses 15\% fewer shell steps and 14\% less model cost, with both policies resolving every issue.}
We evaluate a fixed sequence of 19 issues from \texttt{tablib} and \texttt{tenacity} using Continual Learning Bench~\cite{asawa2026continuallearningbench}, with five runs per arm. Both arms reflect after each graded issue; there is no separate held-out phase. Exo averages 6.28 shell steps and \$0.436 per issue, versus 7.42 and \$0.507 for the control, including reflection. Its observed self-changes are tool and skill edits between issues.

\vspace{1em}
\para{Limits of adaptation.}
Access to self-editing provides a benefit only when the model is capable enough to discover and reuse effective changes. In additional experimentation running Exo on AppWorld with \texttt{gpt-5-mini}, three seeds made limited harness edits and did not reuse their created tools on held-out tasks. Total cost and pass rates were similar (\$6.63 and 59\% vs \$6.73 and 55\% for Exo vs control). Thus, we believe there is an intelligence floor necessary to reap the benefits of Exo's architecture.

\FloatBarrier

\section{Related Work}
\label{sec:related-work}

\para{Learning from execution feedback.}
Execution feedback can guide future behavior through retained lessons~\citep{shinn2023reflexion,zhao2024expel,ouyang2025reasoningbank} or revisions to prompts and workflows~\citep{zhang2025ace,agrawal2025gepa,yuksekgonul2024textgrad,cheng2024trace}. The Self-Defining Operator focuses on incident response, learning reusable playbooks and detectors~\citep{li2026sdo}. In Exo, the running agent can also revise the code governing when and how it learns from feedback.

\para{Harness construction and self-modification.}
Harness search automates policy generation and refinement~\citep{lee2026metaharness,lou2026autoharness,xia2026rrsi}, including evolution over unlabeled test batches~\citep{nie2026tthe}. JIT-Agent generates task-conditioned harnesses with immutable history and mutable controller state~\citep{zhang2026jitagent}. Self-modification spans changes to running code~\citep{yin2024godelagent}, benchmark-driven agent evolution~\citep{zhang2025dgm}, and editable improvement procedures~\citep{zhang2026hyperagents}. Concurrent work, Harness$^2$, refines task harnesses without ground-truth verifiers and updates an improvement harness across tasks~\citep{xu2026harness2}. Exo couples self-modification to task continuation: an accepted revision continues the task, retaining its history and environment.

Within-task adaptation differs in what remains fixed. Live-SWE-agent creates tools while retaining its underlying agent loop~\citep{xia2025liveswe}; Continual Harness periodically invokes a refiner over prompts, sub-agents, skills, and memory~\citep{karten2026continualharness}. Exo also exposes the executor code governing context construction, model calls, and refinement. The acting agent can change both these behaviors and the policy deciding when and how to revise them.

\para{Runtime support for adaptation.}
Runtime adaptation requires both extensibility and independent control over state and execution. DeepSeek Harness combines replaceable plugins with an append-only event log~\citep{deepseek2026harness}, using Cordis for hot replacement and reversible component effects~\citep{shi2026cordis}. Independent checks govern which actions and changes take effect: LogAct couples a shared log with decoupled voters admitting task actions~\citep{balakrishnan2026logact}, while Self-Healing Harness admits behavioral rules through replay or forward trials~\citep{tayebati2026selfhealing}. Exo uses this separation to let the agent replace the policy governing an ongoing task while keeping execution history and replacement authority outside that policy's control.

\section{Discussion}
\label{sec:discussion}

\para{Building bitter-lesson-aligned systems.}
\label{sec:discussion-os}
As reasoning capabilities grow, we believe it is important that the systems we build around them provide benefits that persist and scale proportionally with that growth, rather than being obviated as per the Bitter Lesson~\cite{sutton2019bitterlesson}. Doing so requires a reprioritization of traditional design goals: \emph{flexibility}, \emph{modifiability}, and \emph{legibility} must be elevated to first-class design principles. In the limit, this suggests placing all harness policy, including how it improves itself, under model control, retaining only the mechanisms needed for safe self-evolution. Exo pursues this design with a minimal protected core and an editable, inspectable policy layer. Notably, this does not prescribe a "thin" harness: the agent can and does build substantial domain-specific tooling or remove machinery it no longer needs. The appropriate harness can change with both the model and its workload.

\para{Runtime legibility enables better decision making.} In order to make self-improving systems effective, they need visibility into their own behavior. Policy source code exposes what can be changed, while queryable history records what happened and cost, latency, and cache annotations expose resources consumed. These observations provide denser reward signals about execution efficiency, helping the agent assess and guide improvements before task completion. We believe this to be a key reason for Exo's effectiveness. Preserving history across rollbacks likewise keeps evidence of failed attempts available to future decisions.

\para{A new timescale of system development.}
Historically, the time to build a system has been longer than the timescale of an individual task. As model-assisted development has become faster and task horizons grow, this assumption has begun to flip, and constructing useful specialization can now take less time than the single task that motivates it. Thus, we must increasingly consider development as something that can continue during execution, with procedures synthesized just in time to address the workload at hand. Exo is motivated by this trend, and we see this occur in its evaluations on long-running open-world tasks (\cref{sec:eval-open-world}), with investment in tooling and policy edits paying back before the objective is complete (\cref{sec:eval-combat}). Learning in the deployment environment also grounds revisions in the conditions where they will be used, reducing dependence on representative benchmarks and faithful emulation. This suggests a broader role for continually adapting systems, with trust in their revisions becoming a central systems challenge.

\para{Generalizing guardians.}
\label{sec:discussion-guardians}
Exo's guardian separates the authority to propose a revision from the authority to activate it. The minimal check we provide as default in our implementation is build success; richer guardians could additionally do closer LLM-as-a-judge policy inspection, or even trials in forked environments. The agent could propose additional tests as failures emerge, while changes to the governing acceptance criteria remain independently authorized. More extensive validation incurs additional cost. An open question is how to match validation effort to a revision's consequences while retaining an independent enforcement boundary. We leave exploring additional guardian mechanisms to future work.

\para{Interpretable continual learning.}
\label{sec:discussion-interpretability}
Exo's adaptations appear as explicit changes to code, prompts, and skills. Compared with continual learning encoded solely in model weights, these artifacts expose the learned procedures directly, allowing agent operators to inspect, test, and revise them. Changes can also be examined alongside the execution history that motivated them. This is particularly valuable in domains where operators need to understand, audit, and authorize changes to system behavior. Future guardians could present proposed revisions, supporting evidence, and validation results to a human for approval or rejection before activation. This would let organizations selectively adopt learned procedures through existing review processes.

\para{Exo as an agent runtime.}
\label{sec:discussion-scaling}
Exo's agent runtime decomposition into executor and protected state provides additional benefits in a practical deployment setting. Agents can switch between executor implementations while retaining their durable history and artifacts. Shared runtime services support multiple agents executing concurrently on a single node, each with its own policy and state. Sandbox execution can be offloaded to remote providers, while interchangeable environment backends let deployments change where and how task actions execute without rewriting agent policy. These capabilities allow executor hosting, persistence, and environment resources to be provisioned separately. We leave reporting on a systematic evaluation of agent density, throughput, and resource efficiency to future work.

\para{Future work.}
\label{sec:discussion-future}
A natural next step from Exo's current architecture is exploring parallel self-improvement through explicit agent lineages: fork canonical state, record the parent--child relationship in the event log, and provide a communication channel between branches. Agents could then explore revisions concurrently and share findings with traceable provenance. We leave this to future work.
\section{Conclusion}
\label{sec:conclusion}

We presented Exo, an agent runtime architecture for in-harness learning and adaptation during ongoing work. Exo collapses the optimization loop into task execution, allowing agents to revise all aspects of their harness at runtime. Protected mechanisms preserve execution history and govern executor replacement, while learned adaptations remain explicit and inspectable. Across open-world objectives and graded benchmarks, our evaluation demonstrates efficiency gains during ongoing work and reuse of learned procedures on subsequent tasks. Exo offers a path toward systems whose policies can evolve with their workloads and the capabilities of their underlying models.

\bibliographystyle{mlsys2025}
\bibliography{reference}

\section*{AI Usage Disclosure}
The authors used Claude Code and Codex to assist with inspecting and organizing experiment results and some figure rendering. All coding was done collaboratively with Claude Code or Codex with models varying from April 2026 through July 2026, with human-written design documents and collaborative programming. Paper writing was done primarily by the authors, using codex to assist in revising passages for clarity or brevity, and formatting.

\ifincludeappendix
  \clearpage
  \appendix
\setcounter{figure}{0}
\setcounter{table}{0}
\setcounter{equation}{0}
\renewcommand{\thefigure}{A\arabic{figure}}
\renewcommand{\thetable}{A\arabic{table}}
\renewcommand{\theHfigure}{appendix.\arabic{figure}}
\renewcommand{\theHtable}{appendix.\arabic{table}}
\renewcommand{\theHequation}{appendix.\arabic{equation}}

\section{Preservation Proof and Scope}
\label{supp-app:contracts}

We use the main paper's state $s_t=(d_t,e_t,\pi_t)$, protected transition mechanism $K$, and guardian $G$.

\para{Preservation under self-editing.}
If the initial state satisfies the storage contract, every operation affecting protected state is mediated by $K$, and every permitted transition preserves the contract, then it holds after any finite sequence of operations and executor replacements.

The proof is induction on transitions. The initial state satisfies the contract by assumption. If it holds at step $t$, applying $K$ preserves append-only history and prior artifact versions at step $t+1$, including when the operation replaces $\pi_t$. Induction establishes preservation for any finite execution, independently of how the executor chooses edits.

Similarly, if $G(s,\pi')=1$ implies a candidate property $C(\pi')$ and the initial executor satisfies $C$, every installed executor satisfies $C$ at installation: a transition either retains the current implementation or installs one accepted by $G$.

\para{Enforcement assumptions and limits.}
The storage and replacement mechanisms must remain outside the executor's editing authority. Direct storage access or unrestricted administrative deletion would invalidate complete mediation; the broadly writable development configuration does not enforce these restrictions. Guardian acceptance establishes only the properties actually checked: build success alone guarantees neither task correctness nor improvement. Recovery from a hung executor requires independent authority to stop it and activate a fallback; preservation alone does not ensure liveness. Scoped storage keeps credentials outside sandbox files, but model-context exposure depends on authorized reads and executor tools. We provide such authorized reads, but do not describe their implementation here as it is out of scope for this paper.

\para{Recovery scope.}
Policy and sandbox rollback preserve canonical history and record the rollback. Snapshot scope depends on the provider: Docker image snapshots omit process memory and host-mounted source. Neither reverses external effects, which require domain-specific recovery.

\FloatBarrier
\ifdefined\ifincludeappendix\newpage\fi
\section{Open-World Measurement Details}
\label{supp-app:evaluation}

\Cref{supp-tab:eval-open-world} reports one standalone/transfer pair per objective and harness. Completion costs use the first recorded endpoint at or above the target. Money-making uses the last sampled turn before eight active hours, excluding logged outages.

\begin{table}[ht]
\centering\small
\setlength{\tabcolsep}{2pt}
\definecolor{outcomegood}{HTML}{D8EBD3}
\definecolor{outcomemid}{HTML}{FFE2B7}
\definecolor{outcomebad}{HTML}{F3C6C3}
\caption{Open-world outcomes, one run per entry. Combat/crafting report completion cost; money-making is sampled before eight active hours. Shading compares harnesses within each row and run group: green (best), orange (intermediate), red (worst).}
\label{supp-tab:eval-open-world}
\begin{tabularx}{\columnwidth}{@{}l*{3}{>{\centering\arraybackslash}X}@{\hspace{6pt}}*{3}{>{\centering\arraybackslash}X}}
\toprule
 & \multicolumn{3}{c}{\cellcolor{black!7}\textbf{Standalone}} & \multicolumn{3}{c}{\cellcolor{black!7}\textbf{Transfer}} \\
\cmidrule(lr){2-4}\cmidrule(l){5-7}
Metric & Exo & Codex & Pi & Exo & Codex & Pi \\
\midrule
Combat (\$) $\downarrow$ & \cellcolor{outcomegood}\textbf{3.71} & \cellcolor{outcomebad!68!outcomemid}7.26 & \cellcolor{outcomebad}7.95 & \cellcolor{outcomegood}\textbf{1.92} & \cellcolor{outcomebad!10!outcomemid}6.87 & \cellcolor{outcomebad}10.89 \\
Crafting (\$) $\downarrow$ & \cellcolor{outcomegood}\textbf{2.99} & \cellcolor{outcomebad}3.90 & \cellcolor{outcomemid!92!outcomegood}3.41 & \cellcolor{outcomegood}\textbf{0.88} & \cellcolor{outcomebad}1.89 & \cellcolor{outcomebad!45!outcomemid}1.61 \\
Money (\$) $\downarrow$ & \cellcolor{outcomegood}\textbf{2.84} & \cellcolor{outcomebad}5.14 & \cellcolor{outcomebad!95!outcomemid}5.08 & \cellcolor{outcomegood}\textbf{2.84} & \cellcolor{outcomebad!35!outcomemid}3.68 & \cellcolor{outcomebad}4.08 \\
\quad Coins/\$ (k) $\uparrow$ & \cellcolor{outcomegood}\textbf{332} & \cellcolor{outcomebad!10!outcomemid}231 & \cellcolor{outcomebad}148 & \cellcolor{outcomegood}\textbf{334} & \cellcolor{outcomebad!11!outcomemid}265 & \cellcolor{outcomebad}210 \\
\bottomrule
\end{tabularx}
\end{table}

\fi

\end{document}